\documentclass{article}

\usepackage{microtype}
\usepackage{booktabs} 

\usepackage{hyperref}
\usepackage{algorithm}
\usepackage[algo2e]{algorithm2e}
\usepackage{float}
\usepackage{subfig}
\usepackage{enumitem}
\usepackage{multirow}
\usepackage{arydshln}
\usepackage{wrapfig}
\usepackage[dvipsnames]{xcolor}

\usepackage[accepted]{icml2023}

\usepackage{amsmath}
\usepackage{amssymb}
\usepackage{mathtools}
\usepackage{amsthm}

\usepackage[capitalize,noabbrev]{cleveref}

\theoremstyle{plain}

\theoremstyle{definition}

\theoremstyle{remark}

\DeclareMathOperator*{\argmin}{arg\,min}

\usepackage[textsize=tiny]{todonotes}

\icmltitlerunning{Do Perceptually Aligned Gradients Imply Robustness?}

\begin{document}

\twocolumn[
\icmltitle{Do Perceptually Aligned Gradients Imply Robustness?}



\icmlsetsymbol{equal}{*}

\begin{icmlauthorlist}
\icmlauthor{Roy Ganz}{yyy}
\icmlauthor{Bahjat Kawar}{sch}
\icmlauthor{Michael Elad}{sch}
\end{icmlauthorlist}

\icmlaffiliation{yyy}{Electrical Engineering Department, Technion, Haifa, Israel}
\icmlaffiliation{sch}{Computer Science Department, Technion, Haifa, Israel}

\icmlcorrespondingauthor{Roy Ganz}{ganz@campus.technion.ac.il}

\icmlkeywords{Adversarial Learning, Perceptually Aligned Gradients, ICML}

\vskip 0.3in
]

\newcommand{\ganz}[1]{{\color{red} [ganz: #1]}}



\printAffiliationsAndNotice{}  

\begin{abstract}
Adversarially robust classifiers possess a trait that non-robust models do not -- Perceptually Aligned Gradients (PAG). Their gradients with respect to the input align well with human perception. Several works have identified PAG as a byproduct of robust training, but none have considered it as a standalone phenomenon nor studied its own implications. In this work, we focus on this trait and test whether \emph{Perceptually Aligned Gradients imply Robustness}. To this end, we develop a novel objective to directly promote PAG in training classifiers and examine whether models with such gradients are more robust to adversarial attacks. Extensive experiments on multiple datasets and architectures validate that models with aligned gradients exhibit significant robustness, exposing the surprising bidirectional connection between PAG and robustness.
Lastly, we show that better gradient alignment leads to increased robustness and harness this observation to boost the robustness of existing adversarial training techniques.
Our code is available at \url{https://github.com/royg27/PAG-ROB}.
\end{abstract}

\section{Introduction}
Since the tremendous success of AlexNet~\cite {alexnet}, one of the first Deep Neural Networks (DNNs), in the ImageNet \cite{imagenet} classification challenge, the amount of interest and resources invested in the deep learning (DL) field has skyrocketed. Nowadays, such models attain superhuman performance in classification~\cite{resnet,vit}.
However, although neural networks are 
inspired by the human brain, unlike the human visual system, they are known to be highly sensitive to minor corruptions \citep{corrupt1,corrupt2,corruption3,corruption4,corruption5,corruption6} and small malicious perturbations, known as adversarial attacks \citep{szegedy2014intriguing,synthesizing_robust,Evasion_Attacks,not_easily,goodfellow2015explaining,kurakin2017adversarial,Easily_Fooled}.
With the introduction of such models to real-world applications that affect human lives, these issues raise significant safety concerns, and therefore, they have drawn substantial research attention.

\begin{figure}[t!]
    \centering
    \includegraphics[width=\linewidth]{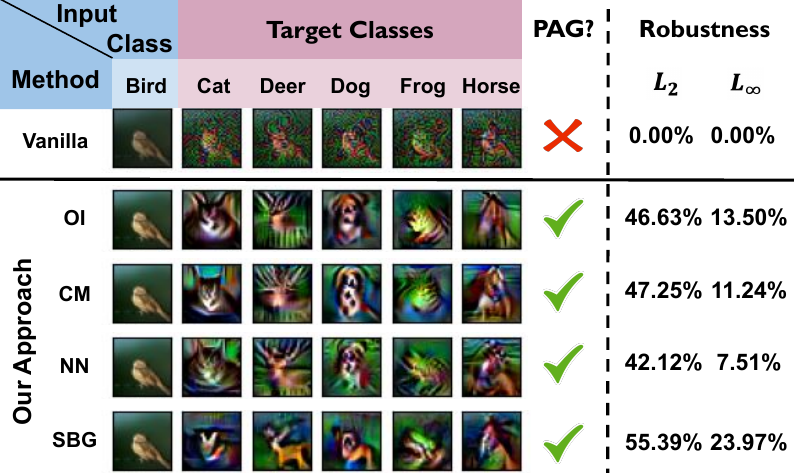}
    \caption{\textbf{PAG implies robustness.} Our method disentangles PAG from adversarial training and allows for a non-adversarial PAG-inducing training scheme that leads to substantial robustness.}
    \label{fig:teaer}
\end{figure}

The bulk of the works in the field of robustness to adversarial attacks can be divided into two types -- on the one hand, ones that propose robustification methods~\citep{goodfellow2015explaining,madry_pgd,trades,mart}, and on the other hand, ones that construct stronger and more challenging adversarial attacks~\citep{goodfellow2015explaining,madry_pgd,carlini_wagner,adaptive_attacks,auto_attack}.
While there are numerous techniques for obtaining adversarially robust models \citep{randomized_smoothing1,randomized_smoothing2,randomized_smoothing3,randomized_smoothing4}, the most effective one is Adversarial Training (AT) \citep{madry_pgd} and its variants~\cite{andriushchenko2020understanding,SAT,PangYDX0020,QinMGKDFDSK19,XieWMYH19,trades,mart}.
AT proposes a simple yet highly beneficial training scheme -- train the network to classify adversarial examples correctly.

While exploring the properties of adversarially trained models, \citet{tsipras2019robustness} exposed a fascinating characteristic of these models that does not exist in standard ones -- Perceptually Aligned Gradients (PAG).
Generally, they discovered that such models are more aligned with human perception than standard ones, in the sense that the loss gradients w.r.t. the input are meaningful and visually understood by humans.
As a result, modifying an image to maximize a conditional probability of some class, estimated by a model with PAG, yields class-related semantic visual features, as can be seen in \cref{fig:pag}.
This important discovery has led to a sequence of works that uncovered conditions in which PAG occurs.
\citet{Aggarwal2020OnTB} revealed that PAG also exists in adversarially trained models with small threat models, while \citet{general_property} observed PAG in robust models trained without adversarial training.
While it has been established that different variations of robust models lead to perceptually aligned gradients, more research is required to understand this intriguing property better.

In this work, while aiming to shed some light on the PAG phenomenon, we pose the following reversed question -- \emph{Do Perceptually Aligned Gradients Imply Robustness?}
This is an interesting question, as it tests the similarity between neural networks and human vision. Humans are capable of identifying the class-related semantic features and, thus, can describe the modifications that need to be done to an image to change their predictions. That, in turn, makes the human visual system ``robust'', as it is not affected by changes unrelated to the semantic features.
With this insight, we hypothesize that since similar capabilities exist in classifiers with perceptually aligned gradients, they would be inherently more robust.

Methodologically testing this question requires training classifiers to obtain PAG without performing robust training.
However, this is challenging as PAG is known to be a byproduct of robust training, and there are currently no ways to promote this property directly and in isolation.
Thus, to explore our research question, we develop a novel PAG-inducing general objective that penalizes the input-gradients of the classifier without any form of robust training.
In particular, our objective encourages the classifier's gradients to be aligned with ``ground-truth'' ones that possess PAG.
As such, our objective requires access to ``ground-truth'' such gradients which are challenging to obtain.
Thus, we explore both heuristic and principled sources for these gradients.
Our heuristic sources stem from the rationale that PAG should point toward the target class.
In addition, we provide in this work a second, principled approach towards creating such PAG vectors, relying on denoising score matching as used in diffusion models~\cite{song2019generative}. 
Specifically, we develop a theoretically justified approach to extract gradients that uphold PAG from diffusion models and encourage the gradients of the classifier to be aligned with ones distilled from diffusion models. 

To validate our hypothesis, we first verify that our optimization goal indeed yields PAG and sufficiently high accuracy on clean images, then evaluate the robustness of the obtained models and compare them to models trained using standard training (``vanilla'').
Our experiments strongly suggest that models with PAG are inherently more robust than their vanilla counterparts, revealing that directly promoting such a trait can imply robustness to adversarial attacks.
We examine this implication across multiple datasets and architectures and conclude that it generally holds and is not data or model dependant.
Surprisingly, although our primary goal is to shed light upon the connection between PAG and robustness rather than proposing a robustification method, not only does our method yield models with non-trivial robustness, but it also exhibits comparable robustness performance to adversarial training without training on perturbed images.
Interestingly, promoting the gradient alignment of a model using our method outperforms adversarial training in the low-data regime and vision-transformers when trained from scratch.
Moreover, it demonstrates better generalization to large $\epsilon$ attacks than AT.
Thus, our findings can potentially pave the way for standard training methods (\textit{i.e.}, without performing adversarial training) for obtaining robust classifiers.
In addition, we study whether there is a correlation between the level of gradient alignment and robustness and discover that models with more aligned gradients are more robust to adversarial examples. Lastly, we harness this insight and introduce our PAG-inducing objective as an auxiliary loss in standard adversarial training techniques and observe that it significantly improves their robustness. 
In particular, we show that improving the gradient alignment of classifiers trained with AT~\cite{madry_pgd} and TRADES~\cite{trades} improves their robustness to seen and unseen attacks by up to $2.24\%$ and $5.25\%$, respectively. 
To summarize:
\begin{itemize}[leftmargin=*] 
    \item We propose a methodological approach to train classifiers to possess PAG without performing adversarial training.
    \item We show that models with aligned gradients are inherently more robust, exposing the bidirectional connection between PAG and robustness.
    \item We demonstrate that increasing gradient alignment improves robustness and leverage this observation to improve existing robustification methods.
\end{itemize}

\section{Background}

\subsection{Adversarial Examples}
\label{sec:adversarial_examples}
We consider a deep learning-based classifier $f_\theta : \mathbb{R}^M \rightarrow \mathbb{R}^C$, where $M$ is the data dimension and $C$ is the number of classes. 
Adversarial examples are instances designed by an adversary in order to cause a false prediction by $f_\theta$~\citep{synthesizing_robust,Evasion_Attacks,not_easily,goodfellow2015explaining,kurakin2017adversarial,Easily_Fooled, szegedy2014intriguing}.
\citet{szegedy2014intriguing} discovered the existence of such samples and showed that it is possible to cause misclassification of an image with an imperceptible perturbation, which is obtained by maximizing the network's prediction error.
Such samples are crafted by applying modifications from a \emph{threat model} $\Delta$ to real natural images.
Hypothetically, the ``ideal'' threat model should include all the possible label-preserving perturbations, \emph{i.e.}, all the modifications that can be done to an image that will not change a human observer's prediction.
Unfortunately, it is impossible to rigorously define such $\Delta$, and thus, simple relaxations of it are used, the most common of which are the $L_{2}$ and the $L_{\infty}$ $\epsilon$-balls: $\Delta = \{\delta \ : \ \lVert\delta\rVert_{c \in \{2, \infty\}} \leq \epsilon\}$.

More formally, given an input sample $\mathbf{x}$, its ground-truth label $y$ and a threat model $\Delta$, a valid adversarial example $\hat{\mathbf{x}}$ satisfies the following: 
$\hat{\mathbf{x}} = \mathbf{x} + \delta \ s.t. \ \delta \in \Delta, y_{pred}\neq y$, where $y_{pred}$ is the prediction of the classifier on $\hat{\mathbf{x}}$.
The procedure of obtaining such examples is referred to as an \emph{adversarial attack}.
Such attacks can be either untargeted or targeted.
Untargeted attacks generate $\hat{\mathbf{x}}$ to minimize $p_\theta(y|\hat{\mathbf{x}})$, namely, cause a misclassification without a specific target class.
In contrast, targeted attacks aim to craft $\hat{\mathbf{x}}$ in a way that maximizes $p_\theta(\hat{y}|\hat{\mathbf{x}}) \ s.t. \ \hat{y} \neq y$, that is to say, fool the classifier to predict $\hat{\mathbf{x}}$ as a target class $\hat{y}$.



While there are various techniques for generating adversarial examples \citep{goodfellow2015explaining,carlini_wagner,boosting_attacks}, we focus in this work on the Projected Gradient Descent (PGD) method \citep{madry_pgd}. 
PGD is an iterative procedure for obtaining adversarial examples that operates in iterative manner as described in \cref{eq:pgd} below:
\begin{equation}
    \label{eq:pgd}
    Repeat: \delta = Proj_{\epsilon}(\delta + \alpha\nabla_{\delta}\mathcal{L}(f_{\theta}(\mathbf{x}+\delta_{t}), y)).
\end{equation}
$Proj_{\epsilon}$ is a projection operator onto $\Delta$, $\alpha$ is the step size, and $\mathcal{L}(\cdot)$ is the classification loss, usually the cross-entropy:
\begin{equation}
    \label{eq:cross_entropy}
    \mathcal{L}_{CE}(\mathbf{z}, y) = - \log \frac{\exp(\mathbf{z}_y)}{\sum_{i=1}^{C} \exp(\mathbf{z}_i)},
\end{equation}
where $\mathbf{z}_y$ and $\mathbf{z}_i$ are the $y$ and $i$ logits, respectively.
In addition, PGD can be extended to targeted attacks.

\subsection{Adversarial Training}
\label{sec:at}
Adversarial training (AT)~\cite{madry_pgd} is a training procedure that aims to obtain adversarially robust classifiers.
A classifier is adversarially robust if applying small adversarial perturbations to its input does not change its label prediction. 
Adversarial training proposes to obtain such classifiers by solving the following optimization problem:
\begin{equation}
    \label{eq:adv_train}
    \min_{\theta} \sum_{(\mathbf{x},y)\in D} \max_{\delta \in \Delta} \mathcal{L} (f_{\theta}(\mathbf{x}+\delta),y).
\end{equation}

Intuitively, the above optimization trains the classifier to accurately predict the class labels of its hardest perturbed images allowed by the threat model $\Delta$. 
Ideally, $\mathcal{L}$ is the 0-1 loss, \emph{i.e.}, $\mathcal{L}(\mathbf{z},y)=\mathbf{I}(\texttt{argmax}_{i}(\mathbf{z}_i)=y)$ where $\mathbf{I}$ is the indicator function.
Nevertheless, since the 0-1 loss is not differentiable, the cross-entropy loss, defined in \cref{eq:cross_entropy}, is used as a surrogate.
In practice, solving this min-max optimization problem is challenging, and there are several ways to obtain an approximate solution.
The most simple yet effective method is based on approximating the solution of the inner maximization via adversarial attacks, such as PGD \citep{madry_pgd}.
According to this strategy, the above optimization is performed iteratively by first fixing the classifier's parameters $\theta$ and optimizing the perturbation $\delta$ for each example via PGD and then fixing $\delta$ and updating $\theta$.
Repeating these steps results in a robust classifier.
Since its introduction by~\citet{madry_pgd}, various improvements to adversarial training were proposed~\citep{andriushchenko2020understanding,SAT,PangYDX0020,QinMGKDFDSK19,XieWMYH19,trades,mart}, yielding classifiers with improved robustness.

\subsection{Perceptually Aligned Gradients}
\label{sec:pag}
\begin{figure}[ht]
    \centering
    \includegraphics[width=\linewidth]{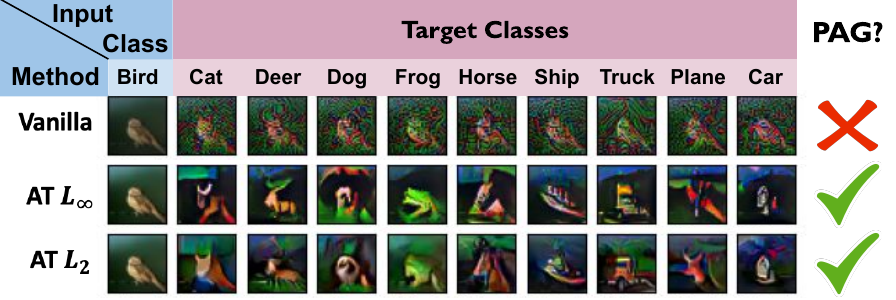}
    \caption{\textbf{Demonstration of PAG property}. Targeted large-$\epsilon$ adversarial examples on robust ResNet-18 trained on CIFAR-10 contain class-related modifications, while "vanilla" ones do not.}
    \label{fig:pag}
    \vspace{-0.5cm}
\end{figure}

Perceptually aligned gradients (PAG) \cite{Engstrom2019AdversarialRA,etmann19a,RossD18,tsipras2019robustness} is a phenomenon according to which classifier input-gradients are semantically aligned with human perception.
It means, inter alia, that modifying an image to maximize a specific class probability should yield visual features that humans associate with the target class.
\citet{tsipras2019robustness} discovered that PAG occurs in adversarially trained classifiers but not in ``vanilla'' ones.
The prevailing hypothesis is that the existence of PAG only in adversarially robust classifiers and not in regular ones indicates that features learned by such models are more aligned with human vision.
PAG is a qualitative trait, and currently, there are no quantitative metrics for assessing it.
Moreover, there is an infinite number of equally good gradients aligned with human perception, \emph{i.e.}, there are countless perceptually meaningful directions in which one can modify an image to look more like a certain target class.
Thus, in this work, similar to \cite{tsipras2019robustness}, we gauge PAG qualitatively by examining the visual modifications done while maximizing the conditional probability of some class, as estimated by the tested classifier.
In other words, we examine the effects of a large-$\epsilon$ targeted adversarial attack and determine that a model has PAG if such a process yields class-related semantic modifications, as demonstrated in \cref{fig:pag}. As can be seen, adversarially trained models obtain PAG, while standardly trained ones (``vanilla'') do not.

In recent years, PAG has drawn a lot of research attention which can be divided into two main types -- an applicative study and a theoretical one. 
The applicative study aims to harness this phenomenon for various computer vision problems, such as image generation and translation~\citep{single_robust}, the improvement of state-of-the-art results in image generation~\citep{ganz2021}, and explainability~\citep{Elliott2021PerceptualBall}.
As for the theoretical study, several works aimed to understand better the conditions under which PAG occurs. 
\citet{general_property} examined if PAG is an artifact of the adversarial training algorithm or a general property of robust classifiers. 
Additionally, it has been shown that PAG exists in adversarially robust models with a low max-perturbation bound~\cite {Aggarwal2020OnTB}.
To conclude, previous works discovered that robust training leads to models with perceptually aligned gradients. In this work, we explore the opposite question -- \emph{Do perceptually aligned gradients imply robustness?}

\section{Do PAG Imply Robustness?}
As mentioned in \cref{sec:pag}, previous work has validated that robust training implies perceptually aligned gradients.
More specifically, they observed that performing targeted PGD attacks on robust models yields visual modifications aligned with human perception.
In contrast, in this work, we aim to delve into the opposite direction and test if training a classifier to have PAG will improve its robustness.

To this end, we propose encouraging the input-gradients of a classifier $f_\theta$ to uphold PAG.
Due to the nature of our research question, we need to disentangle PAG from robust training and verify whether the former implies the latter.
It raises a challenging question -- PAG is known to be a byproduct of robust training. 
How can one develop a training procedure that encourages PAG without explicitly performing robust training of some sort?
Note that a framework that attains PAG via robust training cannot answer our question, as that would involve circular reasoning.
We answer this question by proposing a novel training objective consisting of two elements: the classic cross-entropy loss on the model outputs and an auxiliary loss on the model's input-gradients.
We note that the input-gradients of the classifier, $\nabla_{\mathbf{x}} f_\theta(\mathbf{x})_{y}$, where $f_\theta(\mathbf{x})_{y}$ is the $y$-th entry of the vector $f_\theta(\mathbf{x})$, can be trained, since they are differentiable w.r.t. the classifier parameters $\theta$.
Thus, given labeled images $(\mathbf{x}, y)$ from a dataset $D$, assuming we have access to ground-truth perceptually aligned gradients $g(\mathbf{x}, y_t)$, we could pose the following loss function:
\begin{equation}
    \label{eq:pag_train}
    \begin{split}
    \mathcal{L}_{total}(\mathbf{x}, y) = & \mathcal{L}_{CE}\left(f_{\theta}(\mathbf{x}),y\right) + \\ & \lambda \sum_{y_t=1}^{C}\mathcal{L}_{cos}\left(\nabla_{\mathbf{x}} f_\theta(\mathbf{x})_{y_t}, g(\mathbf{x}, y_t)\right),
    \end{split}
\end{equation}
where $\mathcal{L}_{CE}$ is the cross-entropy loss defined in \cref{eq:cross_entropy}, $\lambda$ is a tunable regularization hyperparameter, $C$ is the number of classes in the dataset, and $\mathcal{L}_{cos}$ is the cosine similarity loss defined as follows:
\begin{equation}
    \label{eq:cosine}
    \mathcal{L}_{cos}(\mathbf{v}, \mathbf{u}) = 1 - \frac{\mathbf{v}^\top \mathbf{u}}{\texttt{max}(\lVert \mathbf{v} \rVert_2 \cdot \lVert \mathbf{u} \rVert_2, \varepsilon)},
\end{equation}
where $\varepsilon$ is a small positive value so as to avoid division by zero.
We emphasize that, in contrast to robust training methods such as AT and randomized smoothing~\cite{madry_pgd,cohen19c}, our scheme does not feed the model with any perturbed images and only trains on examples originating from the training set.
Moreover, while other works \citep{inpput_grads_reg,jakubovitz2018improving} suggest that penalizing the input-gradients' norm yields robustness, we do not utilize this fact since we encourage gradient alignment rather than having a small norm.
Thus, our method is capable of promoting PAG without utilizing robust training of any sort, making it suitable for exploring our titular research question.

After training a model to minimize the objective in \cref{eq:pag_train}, we aim to examine if promoting PAG in a classifier increases adversarial robustness. First, to verify that the resulting model indeed upholds PAG, we perform large-$\epsilon$ targeted PGD on test set images and qualitatively assess the validity of the resulting visual modifications, as in \cite{tsipras2019robustness}.
Afterward, we test the adversarial robustness of the said model and compare it with vanilla baselines. If it demonstrates more aligned gradients and, in return, favorable robustness accuracy, we will have promoted an affirmative answer to the research question of this work. 

However, one major obstacle remains in the way of training this objective: so far, we have assumed the existence of ``ground-truth'' model input-gradients, $g(\cdot,\cdot)$, an assumption that does not hold in practice, as there is no clear way of obtaining point-wise realizations of them.
In the following section, we begin by presenting practical and straightforward heuristic methods for obtaining approximations for these gradients, which we then use for training PAG-promoting classifiers.
Next, we utilize diffusion models for obtaining theoretically justified such gradients as a better source of perceptually aligned gradients.

\section{How are ``Ground Truth'' PAG Obtained?}
\label{sec:grad_sources}

In order to train a classifier for minimizing the objective in \cref{eq:pag_train}, a ``ground truth'' perceptually aligned gradient $g(\mathbf{x}, y_t)$ needs to be provided for each training image $\mathbf{x} \in D$ and for each target class ${y_t \in \{1, 2, \dots, C\}}$.
Since a true such gradient is challenging to obtain, we instead explore a few general pragmatic approaches for approximating these PAGs, beginning with heuristic approaches and then advancing to theoretically justified ones.

\subsection{Target Class Representatives}
\label{sec:simple_method}
\begin{figure}
    \centering
    \includegraphics[width=\linewidth]{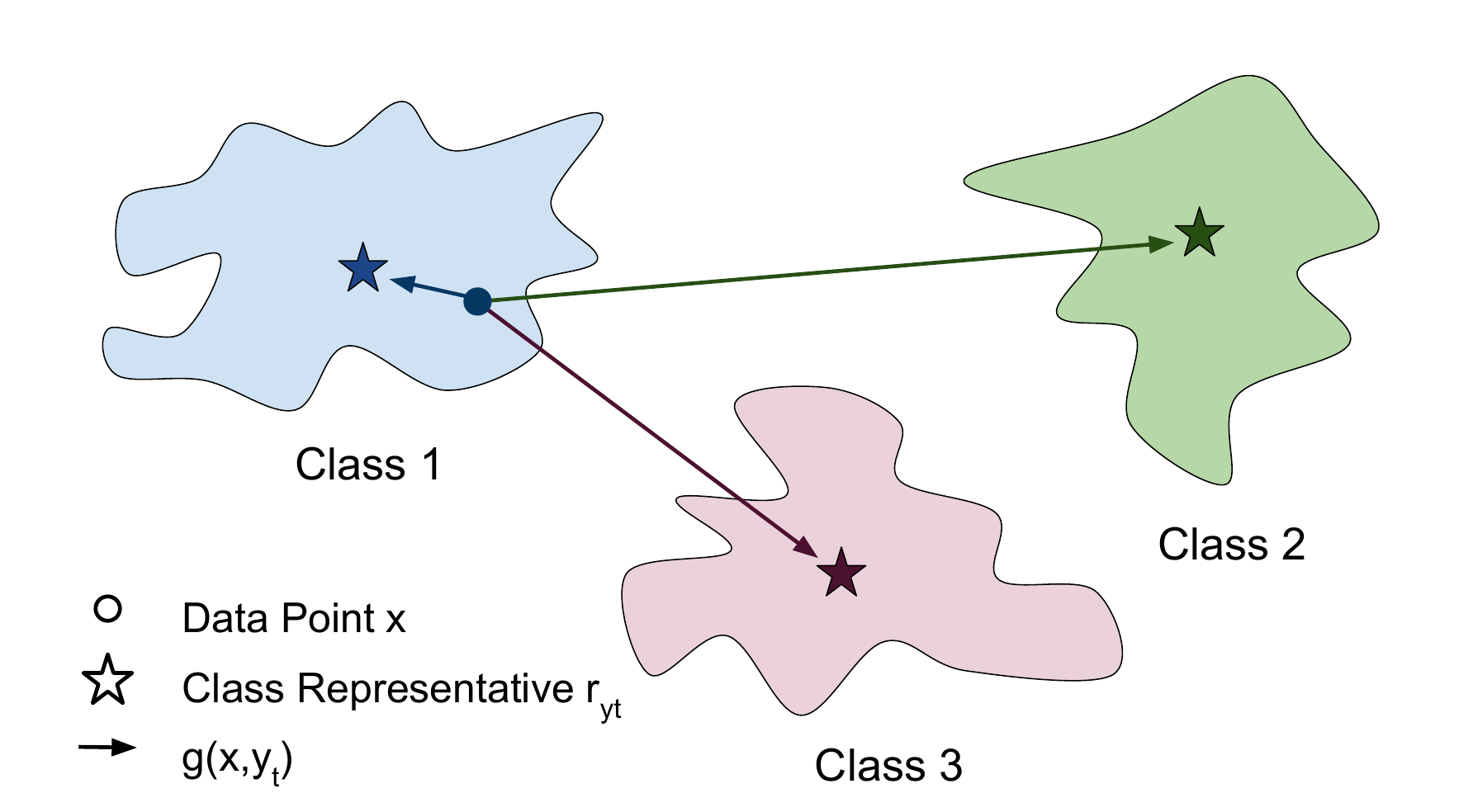}
    \caption{\textbf{Target class representatives}. An illustration of the heuristic realization of perceptually meaningful gradients.}
    \label{fig:class_rep}
\end{figure}

As explained above, we aim to explore ``ground truth'' gradients that promote PAG without relying on robust models.
To this end, we adopt the following simple premise: the gradient $g(\mathbf{x}, y_t)$ should point towards the general direction of images of the target class $y_t$.
Therefore, given a representative of the target class, $\mathbf{r}_{y_t}$, we set the gradient to point away from the current image and towards the representative, \emph{i.e.}, ${g(\mathbf{x}, y_t) = \mathbf{r}_{y_t} - \mathbf{x}}$.
This general heuristic, visualized in \cref{fig:class_rep}, can be manifested in various ways, of which we consider the following:

\noindent\textbf{One Image (OI)}: 
Set $\mathbf{r}_{y_t}$ to be an arbitrary training set image with label $y_t$, and use it as a global destination of $y_t$-targeted gradients.

\noindent\textbf{Class Mean (CM)}: 
Set $\mathbf{r}_{y_t}$ to be the mean of all the training images with label $y_t$. This mean can be multiplied by a constant in order to obtain an image-like norm.

\noindent\textbf{Nearest Neighbor (NN)}: 
For each image $\mathbf{x}$ and each target class $y_t \in \{1,2\dots,C\}$ we set the class representative $\mathbf{r}_{y_t}(\mathbf{x})$ (now dependent on the image) to be the image's NN amongst a limited set of samples from class $y_t$, using $L_2$ distance in the pixel space.
More formally, we define
\begin{equation}
    \mathbf{r}(\mathbf{x},y_t)= \argmin_{\hat{\mathbf{x}}\in D_{y_t} \ \textrm{s.t.} \ 
    \hat{\mathbf{x}}\neq \mathbf{x}}\lVert\hat{\mathbf{x}}-\mathbf{x}\rVert_{2},
\end{equation}
where $D_{y_t}$ is the set of sample images with class $y_t$.

\subsection{Score-Based Gradients}
\label{sec:sbg}
In this section, we describe our principled approach for approximating perceptually aligned gradients via Denoising Diffusion Probabilistic Models (DDPMs), which recently emerged as an interesting generative technique~\citep{sohl2015deep, song2019generative, ho2020denoising}.
These models define noisy versions of an image $\mathbf{x}$, noted as $\left\{\mathbf{x}_t\right\}_{t=1}^T$, and their corresponding noisy data distributions $\left\{p_t(\mathbf{x}_t)\right\}_{t=1}^T$.\footnote{The dependence of $p_t(\cdot)$ on $t$ is omitted to simplify notations.}
Sampling is performed through an iterative process that starts from a Gaussian noise and follows the direction of the \emph{score function}, defined as $\nabla_{\mathbf{x}_t} \log p(\mathbf{x}_t)$ and estimated by a neural network.
Other works~\citep{ho2022cascaded,guided_diffusion} have proposed to provide the class information to such networks, enabling them to model a class-conditional score function $\nabla_{\mathbf{x}_t} \log p(\mathbf{x}_t | y)$.
We observe the similarity between the class-conditional score function and classification loss gradients w.r.t. the input image and hypothesize that gradients distilled from DDPM can be an improved source for perceptually aligned gradients ($g(\cdot,\cdot)$ in \cref{eq:pag_train}).
We factorize the class-conditional score function via Bayes' rule as follows:
\begin{equation}
    \nabla_{\mathbf{x}_t} \log p(\mathbf{x}_t | y) = 
    \nabla_{\mathbf{x}_t} \log p(y | \mathbf{x}_t) +  \nabla_{\mathbf{x}_t} \log p(\mathbf{x}_t),
\end{equation}
leading to
\begin{equation}
    \label{eq:score_based_grad}
    \nabla_{\mathbf{x}_t} \log p(y | \mathbf{x}_t) =
    \nabla_{\mathbf{x}_t} \log p(\mathbf{x}_t | y) - \nabla_{\mathbf{x}_t} \log p(\mathbf{x}_t).
\end{equation}
This equation brings forth a novel usage of diffusion models -- a principled way to estimate the correct gradients for the expression $\log p(y | \mathbf{x}_t)$.
However, classification networks operate on clean images ($\mathbf{x}$) rather than noisy ones ($\mathbf{x}_t$). Therefore, in order to connect classifier input-gradients to DDPMs, we assume that ${\log p(y | \mathbf{x}) \approx \log p(y | \mathbf{x}_t)}$, for specific noise levels $t$.
As a result, the desired estimation for ``ground-truth'' classifier input-gradients can be obtained by subtracting an unconditional score function from a class-condition one.
The choice of $t$ when distilling gradients via this approach introduces a tradeoff -- too large values lead to gradients irrelevant to the input image, while too small ones lead to perceptually meaningless ones (in low noise levels, the conditional and unconditional scores are almost identical). Thus, we set $t$ to be of medium values, yielding both perceptually and image-relevant gradients.
We refer to this technique as Score-Based Gradients (SBG).
We provide additional background and implementation details in \cref{sec:real,sec:ddpm}, respectively.

\section{Experimental Results}
\label{sec:experiments}
In this section, we empirically assess whether promoting PAG in classifiers improves the adversarial robustness at test time.
We experiment using both synthetic and real datasets and present our findings in the following section.
Moreover, we study the correlation between PAG and robustness, \textit{i.e.}, whether more aligned gradients yield improved robustness.

\subsection{A Toy Dataset}
\label{sec:exp_toy}
\begin{figure}[htp] 
    \centering
    \subfloat[Vanilla Training]{%
        \includegraphics[width=0.47\linewidth]{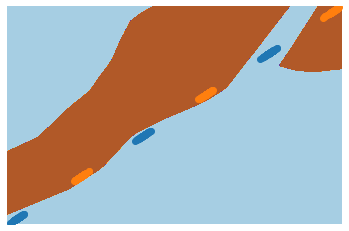}%
        \label{fig:toy_baseline}%
        }%
    \hspace{0.cm}
    \subfloat[Our Training]{%
        \includegraphics[width=0.47\linewidth]{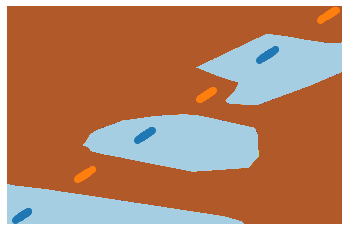}%
        \label{fig:toy_ours}%
        }%
    \caption{\textbf{Decision boundary visualization on the toy dataset}. Visualization of the decision boundary on a synthetic two-class dataset -- the points are the test samples, and the background color represents the predicted class. \cref{fig:toy_baseline,fig:toy_ours} present the decision boundary of a vanilla training method and ours, respectively.}
    \label{fig:toy}
\end{figure}

To illustrate and better understand the proposed approach and its effects, we experiment with a synthetic 2-dimensional dataset and compare our \emph{nearest neighbor} method with the vanilla training scheme that minimizes the cross-entropy loss.
We train a two-layer fully-connected classifier twice: with our nearest neighbor method and without it. We then examine the obtained accuracies and visualize the decision boundaries.
While both methods reach a perfect accuracy over the test set, the obtained decision boundaries differ substantially, as can be seen in \cref{fig:toy}.
The baseline training method results in \cref{fig:toy_baseline} yields dimpled manifolds as decision boundaries, as hypothesized by \citep{dimpled} -- 
the decision boundary of DNN is very close to the data manifold, exposing the model to malicious perturbations.
In contrast, in \cref{fig:toy_ours}, the margin between the data samples and the decision boundary obtained using our approach is significantly larger than the baseline.
This observation helps explain the following robustness result: our model achieves a $75.5\%$ accuracy on a simple adversarial PGD attack, whereas the baseline model collapses to $0.0\%$.
The notion of ``perceptually aligned'' gradients admits a very clear meaning in the context of our 2-dimensional experiment -- faithfulness to the known data manifold.
Therefore, our empirical findings strongly attest that PAG imply robustness in the synthetic use case.
We provide additional experimental details in Appendix \ref{sec:toy}.

\subsection{Real Datasets}
\label{sec:exp_real}
With the encouraging findings presented in \cref{sec:exp_toy}, we now turn to conduct thorough experiments to verify if indeed promoting PAG leads to improved adversarial robustness on real datasets.
To methodologically answer this question, we train models to minimize \cref{eq:pag_train} and follow a two-step evaluation procedure.
First, we validate that models trained with our approach obtain PAG, and next, we test their robustness.
As for examining if a model has PAG, we qualitatively probe whether modifying an image to maximize a certain class probability, estimated by a model, leads to a meaningful class-related change.
For evaluating the robustness, we adopt AutoAttack (AA)~\citep{aa} using both the $L_\infty$ and $L_2$ threat models.
Specifically, we compare models trained to minimize our objective, using the different realizations of PAG proposed in \cref{sec:grad_sources}, with standard (vanilla) classifiers.
Moreover, as a reference point, we also compare the results with adversarially trained (AT)~\citep{madry_pgd} models.
To establish empirical findings beyond a specific dataset and architecture, we experiment with models from different architecture families -- Convolutional Neural Networks and Vision Transformers (ViT)~\citep{vit}, and multiple datasets -- CIFAR-10, STL, and CIFAR-100. In addition, we conduct an extensive architecture ablation in \cref{sec:additional_arch}.

\textbf{CIFAR-10}: 
First, we show in Figure \ref{fig:teaer} and \cref{sec:c10_additional_vis} that while vanilla models do not exhibit semantically meaningful changes, our approach does, as intended.
Surprisingly, although our method is trained to have aligned gradients to some ground truth ones only on the data points, the model generalizes to have meaningful ones beyond these points.
We proceed by quantitatively evaluating the performance on clean and adversarial versions of the test set and report the results in \cref{tab:real_results}.
As for the robustness evaluation, we utilize AutoAttack using $\epsilon=8/255$ for $L_\infty$ and $\epsilon=0.5$ for $L_2$.
While the vanilla baseline is utterly vulnerable to adversarial examples, all the tested PAG-inducing techniques improve the adversarial robustness substantially while maintaining competitive clean accuracies.
It strongly suggests that promoting PAG can improve the classifier's robustness in real image datasets.
A closer inspection of the results indicates that our method performs better in the $L_2$ case over the $L_\infty$ one. 
We hypothesize that this stems from the Euclidean nature of the cosine similarity loss used to penalize the model gradients.
Moreover, while both the heuristic-based and the theoretical-based ``ground-truth'' gradient sources substantially increase the robustness, the latter leads to significantly improved performance. Hence, in the following experiments, we mainly focus on the SBG approach.

Interestingly, the robustification obtained by Score-Based Gradients (SBG) is comparable to AT, without training on adversarial perturbations, potentially setting the foundations for non-adversarial methods for robust training.
In addition, SBG outperforms AT $L_2$ on unseen attacks ($L_\infty$) on both RN-18 and ViT without training on adversarial perturbed images.
Additional fascinating results are obtained on ViT
-- despite their tremendous popularity, ViTs are understudied in the field of adversarial robustness, which still mainly focuses on CNNs.
Similar to the findings of \cite{vit_at}, our results show that AT is significantly less effective for training ViTs from scratch.
On the contrary, SBG outperforms it both in robustness and clean accuracy. 
Specifically, it surpasses the clean accuracy of AT $L_2$ by {\textcolor{Green}{$\mathbf{+12.94\%}$}} and the robust accuracy to $L_2$ attacks by {\textcolor{Green}{$\mathbf{+18.03\%}$}}.
Besides the improved quantitative performance, Figure \ref{fig:vit_pag} in Appendix \ref{sec:additional_vis} shows that AT leads to inferior PAG compared to our method, which further attests to the strong bidirectional connection between PAG and adversarial robustness.
In addition, the clean accuracy obtained by SBG is higher than the ``vanilla'' one ($85.75\%$ compared to $80.51\%$).
We hypothesize that the improved performance of SBG in the ViT case stems from the fact that it serves as a beneficial regularization.
Such a regularization is required as CIFAR-10 is a relatively small dataset, and ViT is highly expressive.

\begin{table}
  \caption{CIFAR-10 results using ResNet-18 and ViT.}
  \centering
  \begin{tabular}{ll ccc}
    \toprule
    \multirow{2}{*}{Arch.} & \multirow{2}{*}{Method} & \multirow{2}{*}{Clean} & AA $L_2$ & AA $L_\infty$\\
    & & & $\epsilon=0.5$ & $\epsilon=8/255$\\
    \midrule
    \multirow{7}{*}{RN-18} & Vanilla & $\mathbf{93.61\%}$ & $00.00\%$ & $00.00\%$\\
    & OI & $79.46\%$ & $46.63\%$ & $13.50\%$ \\
    & CM & $81.41\%$ & $47.25\%$ & $11.24\%$ \\
    & NN & $80.65\%$ & $42.12\%$ & $07.51\%$ \\
    & SBG & $78.56\%$& $55.39\%$& \underline{$23.97\%$}\\ 
    \cdashline{2-5}
    & AT ${L_{\infty}}$ & $82.49\%$ & \underline{$56.57\%$} & $\mathbf{37.59\%}$\\
    & AT ${L_{2}}$ & \underline{$86.79\%$} & $\mathbf{60.82\%}$ &$19.63\%$\\
    \midrule
    \multirow{7}{*}{ViT} & Vanilla & \underline{$80.51\%$}& $00.87\%$ & $00.01\%$ \\
    & OI & $78.06\%$ & $15.47\%$ & $00.21\%$ \\
    & CM & $78.98\%$& $13.73\%$ & $00.17\%$ \\
    & NN & $79.00\%$ & $13.91\%$ & $00.15\%$\\
    & SBG & $\mathbf{85.75\%}$ & $\mathbf{61.02\%}$ & \underline{$22.1\%$}\\ 
    \cdashline{2-5}
    & AT ${L_{\infty}}$ & $62.20\%$ & $42.80\%$& $\mathbf{24.62\%}$ \\
    & AT ${L_{2}}$ & $72.81\%$ & \underline{$42.99\%$} & $08.13\%$ \\
    \bottomrule
  \end{tabular}
  \label{tab:real_results}
\end{table}


\textbf{STL}:
To better validate that the connection between PAG and robustness holds in general, we test our approach on STL~\cite{stl}, which contains images of a higher resolution of $96\times96$ pixels.
Besides its resolution, we choose STL mainly due to its relatively small size -- $5,000$ training and $8,000$ test images.
While it is known that low data regimes are Adversarial Training's Achilles' heel, as it requires more training data~\citep{more_data,more_data2}, we aim to investigate the connection between PAG and robustness in such a challenging setup.
We conduct similar experiments with ResNet-18 as in CIFAR-10 and consider $\epsilon=4/255$ for $L_\infty$ and $\epsilon=0.5$ for $L_2$ as the threat models.
We summarize our qualitative PAG results in Figure \ref{fig:stl_pag} in \cref{sec:additional_vis} and our quantitative ones in \cref{tab:datasets}.
As can be seen in the results, both the heuristic and SBG approaches yield models substantially more robust than standard training.
Moreover, interestingly, while AT struggles to obtain decent results, both our heuristic and principled approaches significantly outperform it in clean and adversarial accuracy, while SBG is substantially better.
Specifically, SBG outperforms AT in clean accuracy by a staggering {\textcolor{Green}{$\mathbf{+19.8\%}$}} and in $L_2$ robustness by {\textcolor{Green}{$\mathbf{+19.63\%}$}}.

\textbf{CIFAR-100}: 
While both CIFAR-10 and STL contain only $10$ classes, we aim to examine whether our approach is also effective on datasets with more classes.
To this end, we experiment with ResNet-18 on the CIFAR-100 dataset~\cite{c100}, containing $100$ classes. 
To reduce the computational cost, we minimize our objective in ~\ref{eq:pag_train} using a subset of $10$ classes, randomly selected in each training iteration.
As for the robustness settings, we use the same threat models as in CIFAR-10.
As can be seen in \cref{tab:datasets} and similar to the trends in CIFAR-10 and STL, classifiers with PAG are significantly more robust than ``vanilla'' ones, attesting to the bidirectional connection between PAG and robustness.
Moreover, SBG obtains similar performance as AT $L_2$ without performing adversarial training.


\begin{table}
  \caption{Results on additional datasets using ResNet-18.}
  \centering
  \begin{tabular}{cl ccc}
    \toprule
    Dataset & Method & Clean & AA $L_2$ & AA $L_\infty$\\
    \midrule
    \parbox[t]{2mm}{\multirow{5}{*}{\rotatebox[origin=c]{90}{STL}}} & Vanilla & $\mathbf{82.60\%}$ & $00.00\%$ & $00.00\%$ \\
    & CM & $70.66\%$ & \underline{$58.90\%$} & \underline{$33.71\%$} \\
    & SBG & \underline{$74.79\%$} & $\mathbf{65.96\%}$ & $\mathbf{43.53\%}$ \\
    \cdashline{2-5}
    & AT ${L_{\infty}}$ & $54.90\%$ & $46.33\%$ & $28.30\%$ \\
    & AT ${L_{2}}$ & $54.99\%$ & $46.04\%$ & $23.33\%$ \\
    \midrule
    \parbox[t]{2mm}{\multirow{5}{*}{\rotatebox[origin=c]{90}{CIFAR-100}}} & Vanilla & $\mathbf{74.36\%}$ & $00.00\%$ & $00.00\%$ \\
    & CM & \underline{$58.89\%$} & $19.94\%$ & $02.78\%$\\
    & SBG & $55.94\%$ & \underline{$29.25\%$} & \underline{$08.24\%$} \\
    \cdashline{2-5}
    & AT ${L_{\infty}}$ & $52.92\%$ & $26.31\%$ & $\mathbf{14.63\%}$ \\
    & AT ${L_{2}}$ & $58.05\%$ & $\mathbf{30.51\%}$ & $08.03\%$ \\
    \bottomrule
  \end{tabular}
  \label{tab:datasets}
\end{table}

\subsection{Robustness To Large $\epsilon$ Attacks}
\label{sec:larger}
Despite the great success of adversarial training (AT) in obtaining robust models, one of its well-known limitations is their poor generalization to unseen attacks. 
Specifically, as AT is performed using a specific threat model, it performs satisfactorily against attacks from the trained threat model but can be easily circumvented by different types of attacks (\textit{e.g.}, larger $\epsilon$).
This limitation was listed by~\cite{hendrycks2021unsolved,bai2021recent} as one of the major unsolved problems of adversarial training methods.
As our approach does not perform training on adversarial examples, we hypothesize that it can cope better against unseen attacks.
To examine this, we first train a ResNet-18 on CIFAR-10 using our SBG approach, standard (``vanilla''), and adversarial training using both $L_2$ and $L_\infty$ with $\epsilon$ of $0.5$ and $8/255$, respectively. Next, we examine their performance against two types of unseen attacks -- larger $\epsilon$ and different threat models (different norms) and summarize our results in \cref{tab:large_attack}.
As can be seen, our approach is substantially more robust to unseen attacks. Specifically, although SBG obtains $55.39\%$ on $\epsilon=0.5$, compared to $60.82\%$ of AT $L_2$, it significantly outperforms it under $L_2$ attacks using $\epsilon=1.0$. 
Of course, training AT $L_2$ against $\epsilon=1.0$ would result in a more robust model against this threat model, but at a high cost of clean accuracy. Contrary, our method does not train against any attack and shows an impressive generalization to different threat models.
It is surprising that our method generalizes well far beyond the data samples, as it trains on the data points themselves rather than on $\epsilon$ volume as AT.
These results show an additional and intriguing benefit of our proposed approach -- robustness to unseen attack.

\begin{table}
  \caption{Large attacks on the CIFAR-10 dataset using ResNet-18.}
  \centering
  \begin{tabular}{l ccc}
    \toprule
    \multirow{2}{*}{Method} & \multirow{2}{*}{Clean} & AA $L_2$ & AA $L_\infty$\\
    & & $\epsilon=1$ & $\epsilon=16/255$\\
    \midrule
    Vanilla & $\mathbf{93.61\%}$ & $00.00\%$ & $00.00\%$ \\
    SBG & $78.56\%$ & $\mathbf{30.43\%}$ & \underline{$01.92\%$}\\ 
    \hdashline
    AT ${L_{\infty}}$ & $82.49\%$ & \underline{$25.26\%$} & $\mathbf{08.02\%}$ \\
    AT ${L_{2}}$ & \underline{$86.79\%$} & $18.81\%$ & $00.50\%$ \\
    \bottomrule
  \end{tabular}
  \label{tab:large_attack}
\vspace{-0.2cm}
\end{table}



\subsection{On The Correlation Between PAG and Robustness}
\label{sec:correlation}
\begin{figure}[ht]
    \centering
    \includegraphics[width=\linewidth]{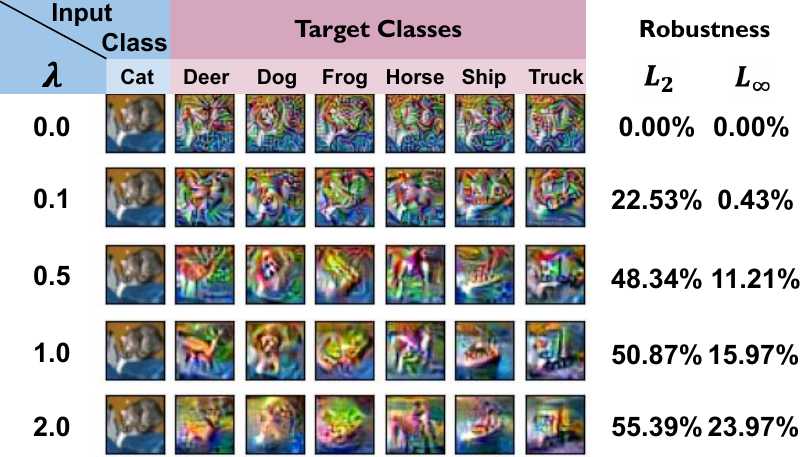}
    \caption{\textbf{Correlation between PAG and robustness}. Demonstration of the positive correlation between PAG and robustness. Higher $\lambda$ values lead to more PAG and in return, more robustness.}
    \label{fig:pag_dose}
\vspace{-0.3cm}
\end{figure}

In previous sections, we discover that models that possess PAG are inherently more robust than ones that do not. 
Our framework enables us to examine whether there is a correlation between the level of the classifier's gradient alignment and its robustness. 
In particular, do models with more aligned gradients more robust?
To empirically answer this question, we train classifiers using SBG with different values of $\lambda$ (\textit{i.e.}, PAG regularization hyperparameter) and conclude our results in \cref{fig:pag_dose}. 
As can be seen, increasing $\lambda$ yields models with more aligned gradients, as the modified images are more semantically similar to the different target classes. 
Interestingly, such models are also significantly more robust than the ones with less aligned gradients.
These findings attest that PAG and robustness are highly correlated, and models with more PAG are substantially more robust.

\section{Improved AT Via Gradient Alignment}
\label{sec:imp_at}
The results in \cref{sec:correlation} show that increasing the level of PAG results in more robust models.
As adversarial training methods are known to yield models that possess PAG, it brings the question of whether such methods can be further improved by increasing the level of gradient alignment.
Specifically, this can be explored by introducing our PAG-inducing objective into adversarial training techniques.
In particular, we experiment on CIFAR-10 with AT~\cite{madry_pgd}  using $L_2, \epsilon=0.5$ and TRADES~\cite{trades} using $L_\infty, \epsilon=8/255$, with ResNet-18 and Wide ResNet 34-10, respectively.
Our results (\cref{tab:improved_at}) show that introducing our loss to adversarial training techniques increase their robustness against the trained threat model by \textcolor{Green}{$\mathbf{+0.91\%}$}, \textcolor{Green}{$\mathbf{+2.24\%}$}, for AT $L_2$ and TRADES $L_\infty$, respectively. 
Moreover, it increases, even more, their robustness against the unseen attack by \textcolor{Green}{$\mathbf{+3.89\%}$}, \textcolor{Green}{$\mathbf{+5.25\%}$}, for AT $L_2$ and TRADES $L_\infty$, respectively.
It bodes well with the generalization findings in \cref{sec:larger}.
We provide additional details in \cref{sec:pag_w_at} and robustness evaluation against additional attacks in \cref{tab:additional_attacks}.

\begin{table}
  \caption{Improving AT via gradient alignment.}
  \centering
  \begin{tabular}{l ccc}
    \toprule
    \multirow{2}{*}{Method} & \multirow{2}{*}{Clean} & AA $L_2$ & AA $L_\infty$\\
    & & $\epsilon=0.5$ & $\epsilon=8/255$\\
    \midrule
    AT $L_2$ & $86.79\%$ & $60.82\%$ & $19.63\%$ \\
    + ours & $85.54\%$ & $61.73\%$ & $23.52\%$\\ 
    \quad $\Delta$ & {\color{red}\textbf{-1.25}} & {\color{OliveGreen}\textbf{+0.91}} & {\color{OliveGreen}\textbf{+3.89}} \\
    \midrule
    TRADES $L_\infty$& $83.04\%$ & $57.41\%$ & $43.54\%$ \\
    +ours & $84.59\%$ & $62.66\%$ & $45.78\%$\\
    \quad $\Delta$ & {\color{OliveGreen}\textbf{+1.55}} & {\color{OliveGreen}\textbf{+5.25}} & {\color{OliveGreen}\textbf{+2.24}} \\
    \bottomrule
  \end{tabular}
  \label{tab:improved_at}
\vspace{-0.3cm}
\end{table}

\section{Conclusions and Future Work}
While previous work demonstrates that adversarially robust models uphold the PAG property, in this work, we investigate the reverse question -- \emph{Do Perceptually Aligned Gradients Imply Adversarial Robustness?}
We believe that answering this question sheds additional light on the connection between robust models and PAG.
To empirically show that inducing PAG improves classifier robustness, we develop a novel generic optimization loss for promoting PAG without relying on robust models or adversarial training and test several manifestations of it.
Our findings strongly suggest that promoting PAG leads to more robust models, exposing the bidirectional connection between the two.
Interestingly, our SBG approach achieves comparable robustness to AT and even outperforms it in the following setups -- (i) low data regimes; (ii) ViTs; (iii) large $\epsilon$ attacks.
In addition, we discover the positive correlation between PAG and robustness and harness this insight to improve existing robustification techniques substantially.
We believe that the obtained results can be further improved with better realizations of PAG and a more sophisticated method, and we hope this work will serve as a foundation for such a line of research.

\section{Acknowledgements}
This research was partially supported by the Israel Science Foundation (ISF) under Grant 
335/18,  the Crown Family Foundation Doctoral Fellowship, and the Council for Higher Education - Planning \& Budgeting Committee.

\bibliography{example_paper}
\bibliographystyle{icml2023}

\newpage
\appendix
\onecolumn
\section{Related Work}
\paragraph{Input-Gradient Penalty Approaches}
Recent works have explored properties of input gradients that improve adversarial robustness. The authors of~\cite{jakubovitz2018improving} demonstrate that regularizing the Frobenius norm of a classifier's Jacobian to be small improves robustness. Such a method is equivalent to regularizing the norm of each such gradient to be small, similar to~\cite{ross2018improving, finlay2021scaleable}.
This line of work attests that requiring small gradient norms, regardless of their direction, leads to robustness.
Moreover, none of these works promotes nor exhibits perceptually aligned gradients.
In opposition to the above methods, we penalize the direction of the gradients, regardless of their norm.
By showing that promoting PAG improves robustness, our method can be viewed as an alternative input-gradient loss for improving robustness.
Nevertheless, the main goal of our work is to better study PAG and its connection with adversarial robustness.

\paragraph{Robust Knowledge Distillation Approaches}
The robust input-gradient distillation methods~\cite{teacher_student1,teacher_student2} is an additional related line of work. In these works, a \emph{student} classifier model is trained to have similar gradients to a robust \emph{teacher} model. The similarity is defined as either cosine similarity or the inability of a discriminator network to distinguish between the gradients of the two models. These works differ from our work, both in their loss functions and specific utilization of the teacher model, but essentially, they all demonstrate how distilling knowledge from a robust teacher model can invoke adversarial robustness in the student model. While successful in their respective tasks, these methods are unsuitable for assessing whether perceptually-aligned gradients inherently promote robustness, as they implicitly rely on prior adversarial training.

\section{Denoising Diffusion Probabilistic Models}
\label{sec:ddpm}
Denoising diffusion probabilistic models (DDPMs) are a new fascinating generative approach~\cite {sohl2015deep, song2019generative, ho2020denoising}.
Such methods achieve state-of-the-art results in image generation~\citep{guided_diffusion, song2020score, vahdat2021score}, and were additionally deployed in several downstream tasks such as inverse problems~\citep{kawar2021snips, kawar2022denoising}, image compression~\citep{theis2022lossy}, image segmentation~\citep{amit2021segdiff}, image editing~\citep{liu2021more, avrahami2022blended, kawar2022imagic}, and text-to-image generation~\citep{rombach2022high, ramesh2022hierarchical, saharia2022photorealistic}, among others.

The core idea of these models, also known as score-based generative models, is to start from a random Gaussian noise image $\mathbf{x}_T$ and then iteratively denoise it into a photorealistic image $\mathbf{x}_0$ in a controlled manner.
This process can also be interpreted as an annealed version of Langevin dynamics~\citep{song2019generative}, where each iteration $t \in \{T, T-1, \dots, 1\}$ follows the direction of the \emph{score function}, defined as $\nabla_{\mathbf{x}_t} \log p(\mathbf{x}_t)$, with an additional noise for stochasticity.
Each intermediate image $\mathbf{x}_t$ can be considered a noisy version of a pristine image $\mathbf{x}_0$, with a pre-defined noise level $\sigma_t$.
The score function can be estimated using a neural network trained for mean-squared-error denoising~\cite {stein1981estimation, Miyasawa61, efron2011tweedie}.
This estimation can also be generalized for denoising models conditioned on a class label $y$, obtaining $\nabla_{\mathbf{x}_t} \log p(\mathbf{x}_t | y)$~\citep{ho2022cascaded}.
In this work, we propose a novel usage of diffusion models for approximating perceptually aligned gradients and train classifiers to be aligned with them.
Our experiments show that this principled PAG realization is highly reliable, leading to a substantially increased robustness.



\begin{figure}[ht]
    \centering
    \includegraphics[width=0.7\linewidth]{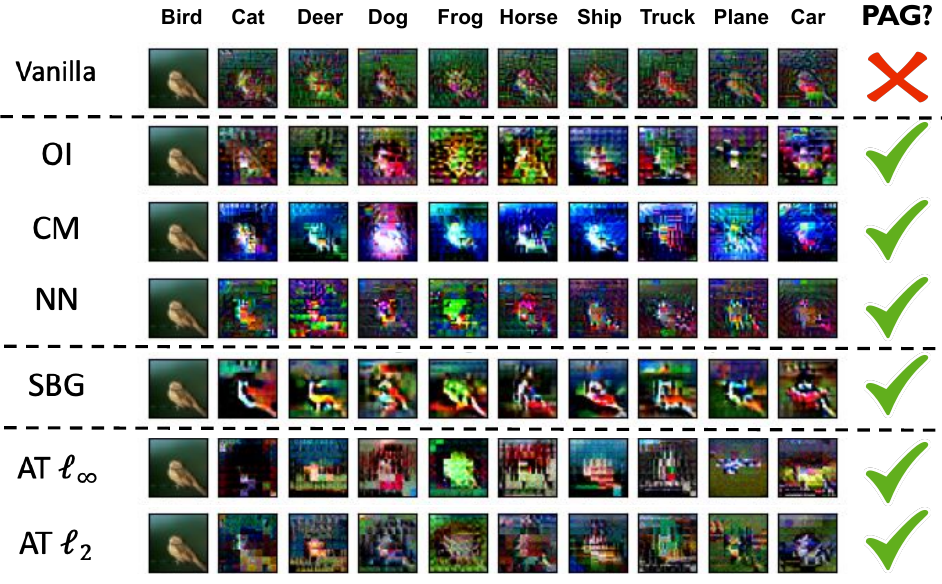}
    \caption{\textbf{Vision Transformer (ViT) PAG examination on CIFAR-10}. Large-$\epsilon$ targeted attacks on ViT trained using standardly (``vanilla''), adversarially (AT), and using both our heuristic (OI, CM, and NN) and principled approaches (SBG) on CIFAR-10.}
    \label{fig:vit_pag}
\end{figure}

\textbf{ViT}: As can be seen in \cref{fig:vit_pag}, while ``vanilla'' classifiers do not possess PAG, both our heuristic approaches and AT models obtain more aligned gradients. Interestingly, our principled SBG approach leads to the best PAG among the tested methods and obtains the highest robustness. 
This further attests to the positive correlation between PAG and robustness.


\subsection{STL}
\begin{figure}[ht]
    \centering
    \includegraphics[width=0.7\linewidth]{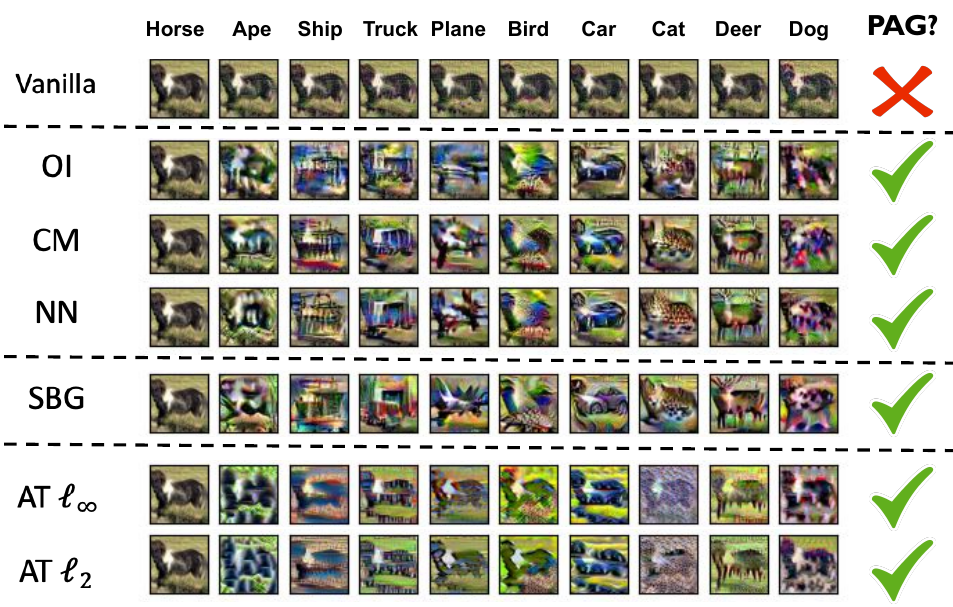}
    \caption{\textbf{ResNet-18 (RN-18) PAG examination on STL}. PAG examination for ViT on CIFAR-10 trained standardly (``vanilla''), adversarially (AT) and using both our heuristic (OI, CM and NN) and principled approaches (SBG).}
    \label{fig:stl_pag}
\end{figure}

\begin{figure}
    \centering
    \includegraphics[width=\linewidth]{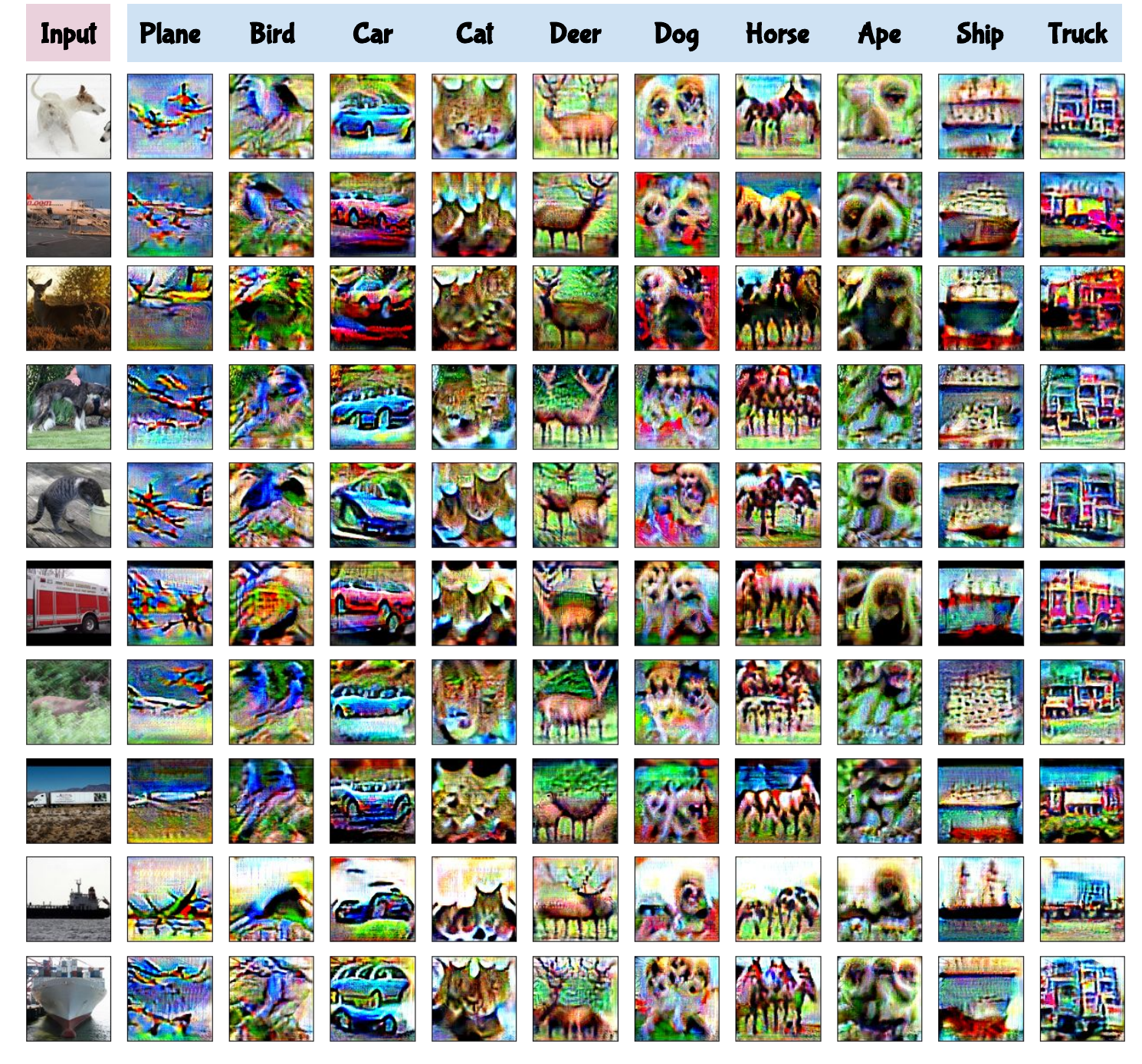}
    \caption{\textbf{STL PAG Visualizations of SBG}. Targeted large-$\epsilon$ adversarial examples on ResNet-18 trained on STL using SBG. We present 10 randomly selected images and transform them into all the possible classes. As can be seen, all images contain meaningful class-related semantic information.}
    \label{fig:stl_pag_rn18_sbg}
\end{figure}

In this section, we qualitatively examine the existence of PAG for ResNet-18 on the STL dataset. 
Specifically, we consider standard-trained classifiers (``vanilla), adversarially trained ones (AT), and models trained using both our heuristic (OI, CM, and NN) and principled (SBG) approaches.
As can be seen in \cref{fig:stl_pag}, while the ``vanilla'' classifier does not possess PAG at all and AT models feature a minimal gradient alignment, our approaches lead to much more aligned gradients.
As in CIFAR-10, we visualize additional demonstrations for the SBG case in \cref{fig:stl_pag_rn18_sbg}.
This further strengthens the connection between PAG and robustness, as our methods achieve much higher robustness than adversarial training on this dataset.

\section{Additional Architectures Ablation}
\label{sec:additional_arch}
In this section, we provide the results of applying our method to additional architecture types.
While we focus in \cref{sec:experiments} on skip-connection-based convolutional NN (ResNet-18) and an attention-based one (ViT), we turn to examine it on other types of architectures.
Specifically, we apply it to VGG~\cite {vgg}, a convolutional network without residual connection, and MLP Mixer~\citep{mlpmixer}, a top-performing dense architecture.
Empirically proving that perceptually aligned gradients imply robustness in these architectures will further strengthen that the connection between the two is model agnostic.
We experiment with such architectures using the SBG approach on the CIFAR-10 dataset and report the results in \cref{tab:more_arch}. We evaluate the performance on clean and adversarially perturbed images using AuttoAttack with a $L_2, \epsilon=0.5$  threat model.
For the MLP Mixer, we follow the CIFAR-10 adjusted implementation\footnote{\url{https://github.com/omihub777/MLP-Mixer-CIFAR}} and train it for 100 epochs using a batch size of $128$. As for our approach, we utilize the SBG PAG realizations and set $\lambda=0.5$.
As for the VGG, we follow the implementation for CIFAR-10\footnote{\url{https://github.com/chengyangfu/pytorch-vgg-cifar10}} and train for 100 epochs using a batch size of $64$. Similarly to MLP Mixer, we use SBG with $\lambda = 0.5$. Moreover, we experiment with VGG-11, VGG-13, and VGG-16 to further study the depth effect. 
As our results suggest, the connection between PAG and robustness is general and architecture-independent.

\begin{table}[ht]
  \caption{Accuracy on CIFAR-10 using VGG and MLP Mixer architectures.}
  \centering
  \begin{tabular}{l c cc}
    \toprule
    Method & Arch. & Clean & AutoAttack $L_2$\\
    \midrule
    Vanilla & \multirow{2}{*}{VGG-16} & $\mathbf{92.32\%}$ & $00.20\%$\\
    SBG & & $81.93\%$ & $\mathbf{42.03\%}$\\
    \hdashline
    Vanilla & \multirow{2}{*}{VGG-13} & $\mathbf{92.47\%}$ & $00.11\%$\\
    SBG & & $82.05\%$ & $\mathbf{41.49\%}$\\
    \hdashline
    Vanilla & \multirow{2}{*}{VGG-11} & $\mathbf{90.82\%}$ & $02.50\%$\\
    SBG & & $79.22\%$ & $\mathbf{35.79\%}$\\
    \midrule
    Vanilla & \multirow{2}{*}{MLP-Mixer} & $\mathbf{72.05\%}$ & $00.50\%$\\
    SBG && $63.04\%$ & $\mathbf{35.97\%}$\\
    \bottomrule
  \end{tabular}
  \label{tab:more_arch}
\end{table}

\section{Robustness to Additional Adversarial Attacks}
In the main paper, we test our approach on $L_2$ and $L_\infty$ norm-bounded adversarial attacks and compare it to Adversarial Training. In this section, we extend the evaluation to additional adversarial attacks, including non-norm bounded ones, using Foolbox\footnote{\url{https://github.com/bethgelab/foolbox}} and report the results in \cref{tab:additional_attacks}. Specifically, we use ResNet-18 trained on CIFAR-10 using SBG and Adversarial Training with $L_2$ threat model and $\epsilon=0.5$.

\begin{table}[ht]
  \caption{Robustness of ResNet-18 trained on CIFAR-10 against various adversarial attacks.}
  \centering
  \begin{tabular}{l cc}
    \toprule
    Attack & AT $L_2$ & SBG\\
    \midrule
    SaltAndPepperNoiseAttack & 29.86\% & \textbf{42.59\%} \\
    BinarySearchContrastReductionAttack & 20.02\% & \textbf{44.71\%}\\
    GaussianBlurAttack & 28.03\% & \textbf{51.54\%}\\
    LinearSearchBlendedUniformNoiseAttack & \textbf{34.95\%} & 34.30\%\\
    L0FMNAttack & \textbf{43.03\%} & 40.96\%\\
    L1FMNAttack & 31.60\% & \textbf{40.24\%}\\
    EADAttack & 8.57\% & \textbf{33.64\%}\\
    \bottomrule
  \end{tabular}
  \label{tab:additional_attacks}
\end{table}

\section{Tiny ImageNet Experiments}
\label{sec:tiny}
To further show that promoting PAG increases adversarial robustness, we conduct experiments on the challenging Tiny ImageNet dataset~\cite{le2015tiny}, containing $100,000$ $64\times64\times3$ images of $200$ classes.
We compare the obtained results of our CM and SBG approaches to the standard training and report the results in \cref{tab:tiny}.
Our results show that models with PAG are inherently more robust.

\begin{table}
  \caption{Results on Tiny ImageNet using ResNet-18.}
  \centering
  \begin{tabular}{l ccc}
    \toprule
    Method & Clean & AA $L_2$ & AA $L_\infty$\\
    \midrule
    Vanilla & $\mathbf{61.19\%}$ & $02.37\%$ & $00.00\%$ \\
    CM & $50.04\%$ & \underline{$19.21\%$} & \underline{$03.14\%$} \\
    SBG & \underline{$57.40\%$} & $\mathbf{25.21\%}$ & $\mathbf{05.50\%}$ \\
    \bottomrule
  \end{tabular}
  \label{tab:tiny}
\end{table}

\section{Visualization of PAG Realizations}
\begin{figure}[ht]
    \centering
    \includegraphics[width=0.7\linewidth]{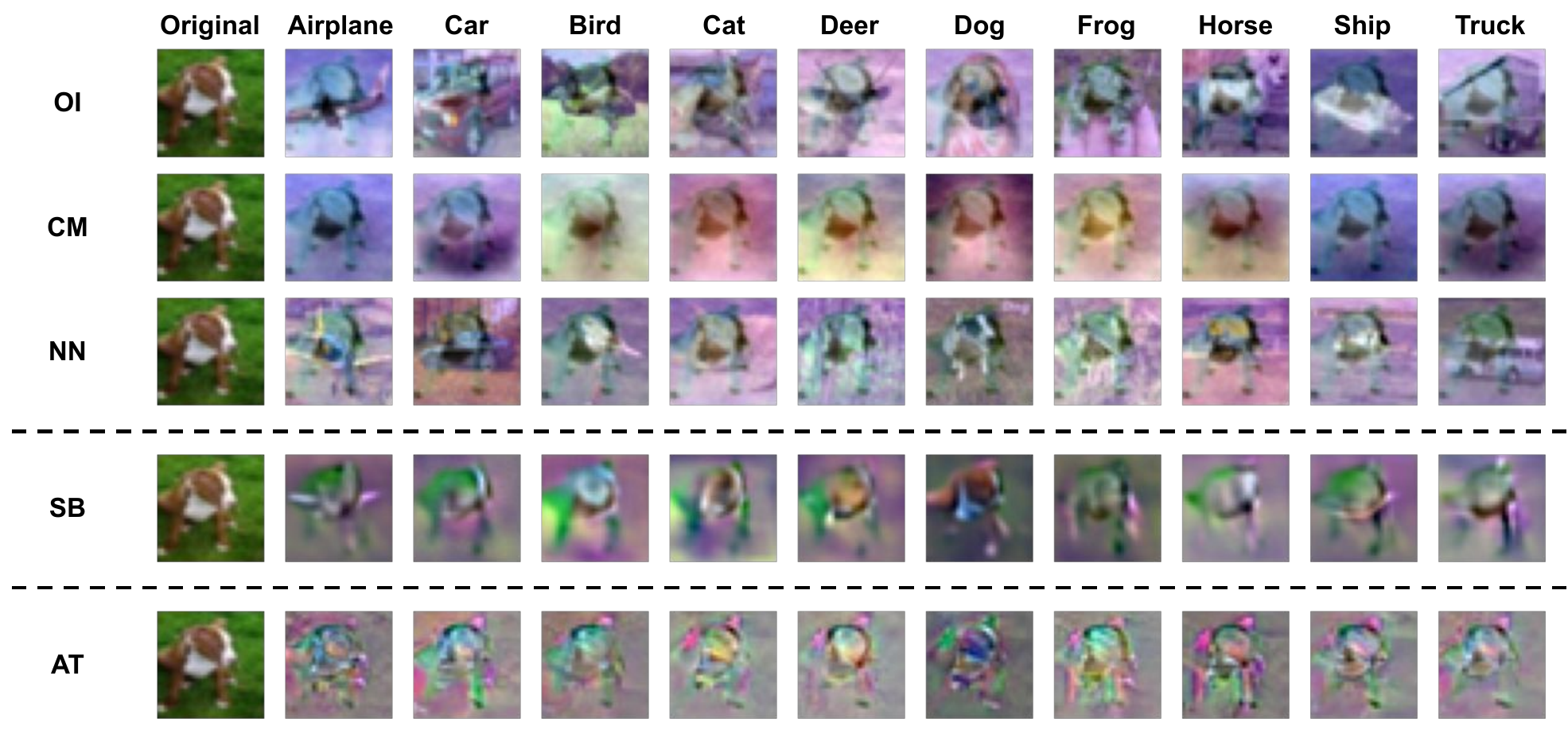}
    \caption{\textbf{Visualization of different realizations of PAG}. We visualize the input-gradients w.r.t. different target classes for the PAG realizations considered in this work -- heuristic (OI, CM, NN) and principled (SBG) realizations. As a reference, we visualize the gradients of the adversarially trained classifier.} 
    \label{fig:gt_grads}
\end{figure}

We visualize our proposed realizations for approximating Perceptually Aligned Gradients in Figure \ref{fig:gt_grads} on CIFAR-10.
In the top three rows, we show the results of the heuristic-based methods.
A ghosting effect can be seen as these gradients derive from the subtraction of two images.
However, in Score-Based Gradients, the modifications focus on the object, and features of the target class can be observed (i.e., horse features ).
The nature of SBG is object-centric as the performed modifications focus on the object itself, similar to AT.

\section{Implementation Details}
\subsection{Toy Dataset}
\label{sec:toy}
\textbf{Data}: We experiment with our approach on a 2-dimensional synthetic dataset to demonstrate its effects.
To this end, we construct a dataset of 6,000 samples from two classes containing precisely 3,000 examples.
Our samples, $\mathbf{x}=[x_1,x_2]$, reside on the straight line $x_2 - 2 x_1 = 0$ in the 2-dimensional space $\mathbb{R}^2$, where each class $y \in \{0, 1\}$ follows a Gaussian mixture distribution.
Each class contains three modes, and each contains 1000 samples drawn from a Gaussian distribution ($x_1\sim N(c,1), x_2=2*x_1$, where $c$ is the mode center). The modes centers are set to be $\{-50,-10,30\}$ and $\{-30,10,50\}$.
This way, the cardinal manifold assumption according to which high-dimensional images reside on a lower-dimensional manifold \cite{low_dim_manifold} holds.
To evaluate performance, we generate a balanced test set from the same distribution consisting of 600 samples.

\textbf{Architecture and Training}: 
We use a 2-layer fully-connected network $(2\rightarrow 32 \rightarrow 2)$ with ReLU non-linearity. We train it twice -- using the standard cross-entropy training and our proposed method with NN realization. 
We do so for $100$ epochs with a batch size of $128$, using Adam optimizer, a learning rate of $0.01$, and the same seed for both training processes.

\textbf{Computational Resources}: We use a single Tesla V100 GPU.

\textbf{Evaluation}: 
As detailed in the paper, we test the performance of the models using standard and adversarial evaluation. We draw 600 test samples from the same distribution as the train set for the standard one and measure the accuracy. As for the adversarial one, we use an $L_2$-based 10-step PGD with $\epsilon=15$ and a step size of $2$. Note that this choice of $\epsilon$ guarantees in our settings that the allowed threat model is too small for actually changing a sample of a certain class to the other one, making it a valid threat model.

\subsection{Real Datasets}
\label{sec:real}
\textbf{Data}:
As for our real datasets experiments, we use CIFAR-10, CIFAR-100, and STL that contain images of size $32\times32\times3$ (the formers) and $96\times96\times3$ (the latter). 
For each dataset and PAG ``ground-truth'' realization, we construct a training set by computing $C$ targeted gradients for each training sample ($C=10$ for CIFAR-10 and STL and $C=100$ for CIFAR-100) for reproducibility and consistency purposes. 

To obtain our Score-Based Gradients (SBG), we follow the implementation of \citep{improved_diff}\footnote{\url{https://github.com/openai/improved-diffusion}} for training a class-conditional diffusion model for CIFAR-10, CIFAR-100,  and STL datasets. We use their CIFAR-10 architecture for the CIFAR datasets and their ImageNet architecture for STL, adapting the image size by a simple bicubic interpolation.
In particular, for a $C$-classes dataset, we train a single class-conditioned diffusion model with $C + 1$ classes, where the additional class represents the absence of class information and thus models the unconditional score function.
Instances of this class are drawn with probability $1/C$, and they originate uniformly from each of the $C$ classes.
This way enables us to distill gradients using \cref{eq:score_based_grad} with the same model, mitigating the need to scale the outputs of the conditional and unconditional models.

\begin{figure}[ht]
    \centering
    \includegraphics[width=0.7\linewidth]{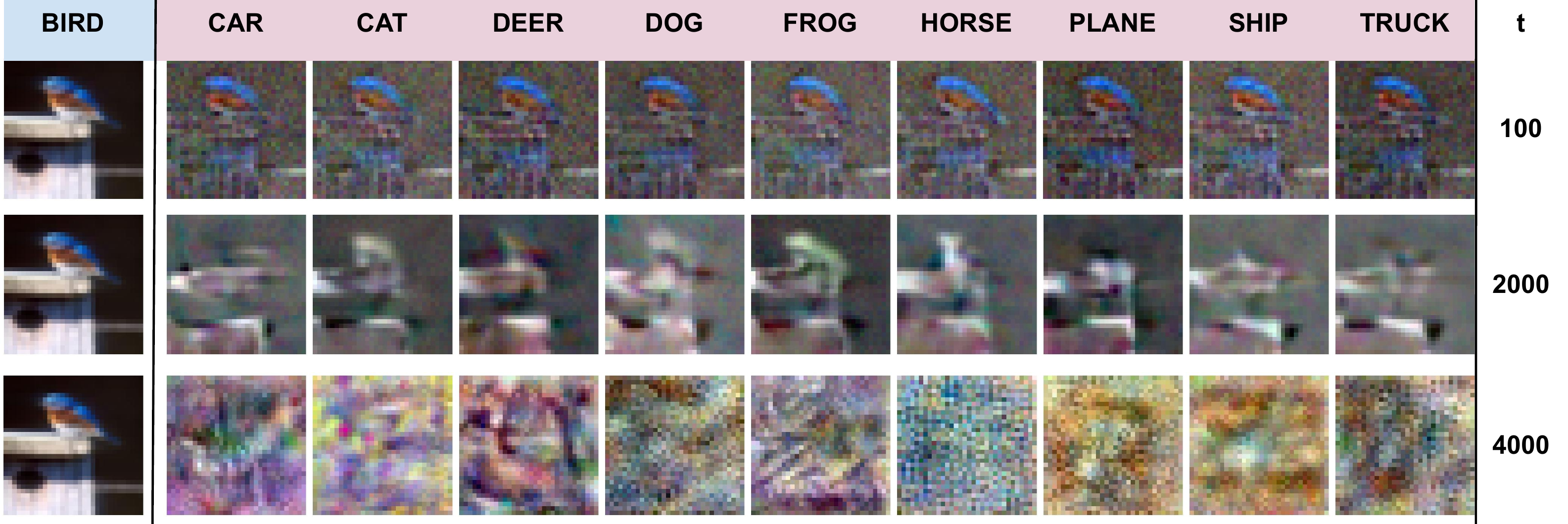}
    \caption{\textbf{Gradients realization using SBG with different $t$ values}. We visualize the effects of different levels of $t$ when distilling gradients using the SBG approach. As can be seen, while too large or small $t$ values lead to meaningless gradients, mid-values of $t$ leads to the desired outputs.} 
    \label{fig:sbg_tradeoff}
\end{figure}

As specified in \cref{sec:sbg}, \cref{eq:score_based_grad} deals with noisy images, not clean ones, as we consider in this work. Nevertheless, picking a proper noise level $t$ enables us to distill meaningful and valuable gradients. As explained in \cref{sec:sbg},  too large values lead to gradients irrelevant to the input image, while too small ones lead to perceptually meaningless ones. However, setting $t$ to an intermediate value leads to semantically meaningful outputs. We demonstrate this in \cref{fig:sbg_tradeoff}.

\textbf{Training}: 
For all the tested datasets, we train the classifier (ResNet-18 or ViT) for $100$ epochs, using SGD with a learning rate of $0.01$, a momentum of $0.9$, and a weight decay of $0.0001$.
In addition, we use the standard augmentations for these datasets -- random cropping with padding of $4$ and random horizontal flipping with a probability of $0.5$.
We use a batch size of $64$ for CIFAR-10 and CIFAR-100 and $32$ for STL.
As for the ViT, we use a publicly available implementation, adjusted to CIFAR-10\footnote{\url{https://github.com/omihub777/ViT-CIFAR}}, containing 6.3 million parameters.
We present in \cref{tab:imp_det} the best choices of $\lambda$ -- the coefficient of our PAG promoting auxiliary loss term in all the tested datasets and methods.
The values of $\lambda$ suggest that higher values should be applied for better gradient sources (\textit{e.g.}, SBG's ideal $\lambda$ is higher than the heuristic methods)

As for our baselines, we use the same training hyperparameters mentioned above. Regarding the AT, we use $7$ steps PGD using a step size of $1.5 * \frac{\epsilon}{7}$ and follow the base implementation presented in \citep{trades}\footnote{\url{https://github.com/yaodongyu/TRADES}} and extend it to $L_2$.

\begin{table}[ht]
  \caption{Values of the hyperparameter $\lambda$.}
  \centering
  \begin{tabular}{l cccc}
    \toprule
    \multirow{2}{*}{Method} & \multicolumn{2}{c}{CIFAR-10} & STL & CIFAR-100\\  &
    RN-18 & ViT & RN-18 & RN-18\\
    \midrule
    One Image &  $0.2$ & $0.1$ & - & - \\
    Class Mean &  $0.2$ & $0.1$ & $0.2$ & $0.5$\\
    Nearest Neighbor &  $0.2$ & $0.1$ & - & - \\
    SBG & $1.5$ & $3$ & $1$ & $1$\\
    \bottomrule
  \end{tabular}
  \label{tab:imp_det}
\end{table}

\textbf{Computational Resources}: We use two NVIDIA RTX A4000 16GB GPUs for each experiment.

\textbf{Evaluation}: We use the de-facto standard evaluation library of AutoAttack~\citep{aa}\footnote{\url{https://github.com/fra31/auto-attack}}.


\subsection{Improved AT Via Gradient Alignment}
\label{sec:pag_w_at}
As demonstrated in \cref{sec:imp_at}, introducing gradient alignment regularization further improves the robustness achieved by robustification methods.
In particular, we show this for Adversarial Training using $L2, \epsilon=0.5$ and TRADES using $L_\infty, \epsilon=8/255$, with ResNet-18 and Wide ResNet 34-10.
We use the official code repository for the latter and implement our loss upon it.
We train each model twice with the same hyperparameters -- with and without our regularization. 
We set $\lambda$ to be $0.2$ for AT and $0.5$ for TRADES.
Next, we used AutoAttack to evaluate the performance of the trained models against both $L_2$ and $L_\infty$ attacks.


\section{Runtime Comparison with Adversarial Training}
\label{app:speed}
As our method does not compute adversarial examples (an iterative process), it is faster than adversarial training.
To quantify this, we conduct a runtime comparison using a batch size of size $1$, using ResNet-18 and the CIFAR-10 dataset, and reveal that our method is faster than AT-PGD-7 and AT-PGD-20 by x2.13 and x6.14, respectively.
However, utilizing our approach with Score-Based-Gradients (SBG) requires access to a trained diffusion model, which takes additional time to train and sample.
Nevertheless, the same diffusion model is utilized to generate the SBG gradients used to train the different classifiers considered in this work.
For example, training a diffusion model on the CIFAR-10 dataset to achieve satisfactory results takes several hours. Nevertheless, as we use a single $t$ value for crafting the SBG, the diffusion training time can be significantly reduced by training only on the corresponding noise level.


\section{Additional PAG Demonstrations}
\label{sec:additional_vis}
\subsection{CIFAR-10}
\label{sec:c10_additional_vis}
In this section, we provide a qualitative examination of the existence of PAG for ResNet-18 and ViT on the CIFAR-10 dataset. 
Specifically, we consider standard-trained classifiers (``vanilla), adversarially trained ones (AT), and models trained using both our heuristic (OI, CM, and NN) and principled (SBG) approaches. 

\begin{figure}[ht]
    \centering
    \includegraphics[width=0.7\linewidth]{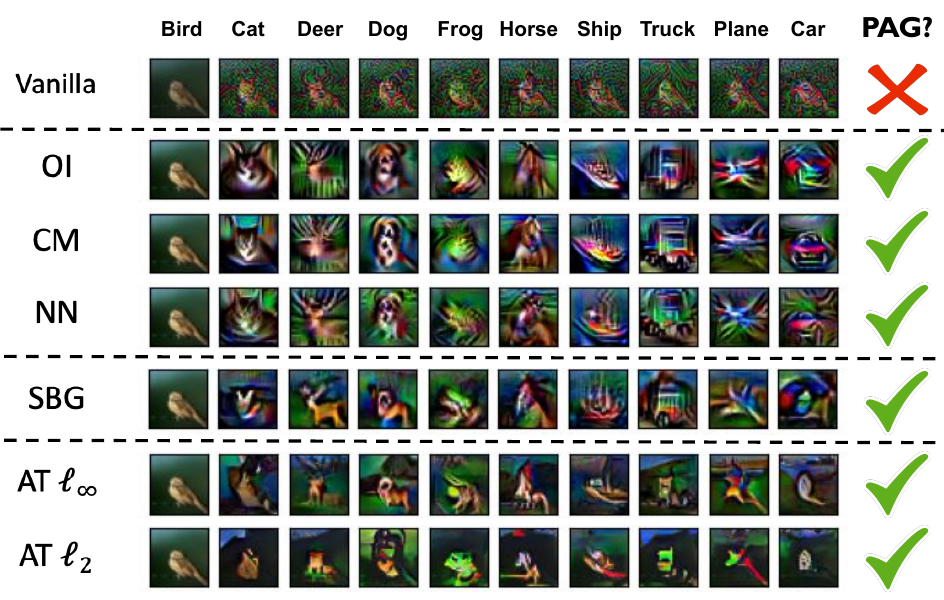}
    \caption{\textbf{ResNet-18 (RN-18) PAG examination on CIFAR-10}. Large-$\epsilon$ targeted attacks on RN-18 trained standardly (``vanilla''), adversarially (AT), and using both our heuristic (OI, CM, and NN) and principled approaches (SBG) on CIFAR-10.}
    \label{fig:rn_pag}
\end{figure}

\begin{figure}
    \centering
    \includegraphics[width=\linewidth]{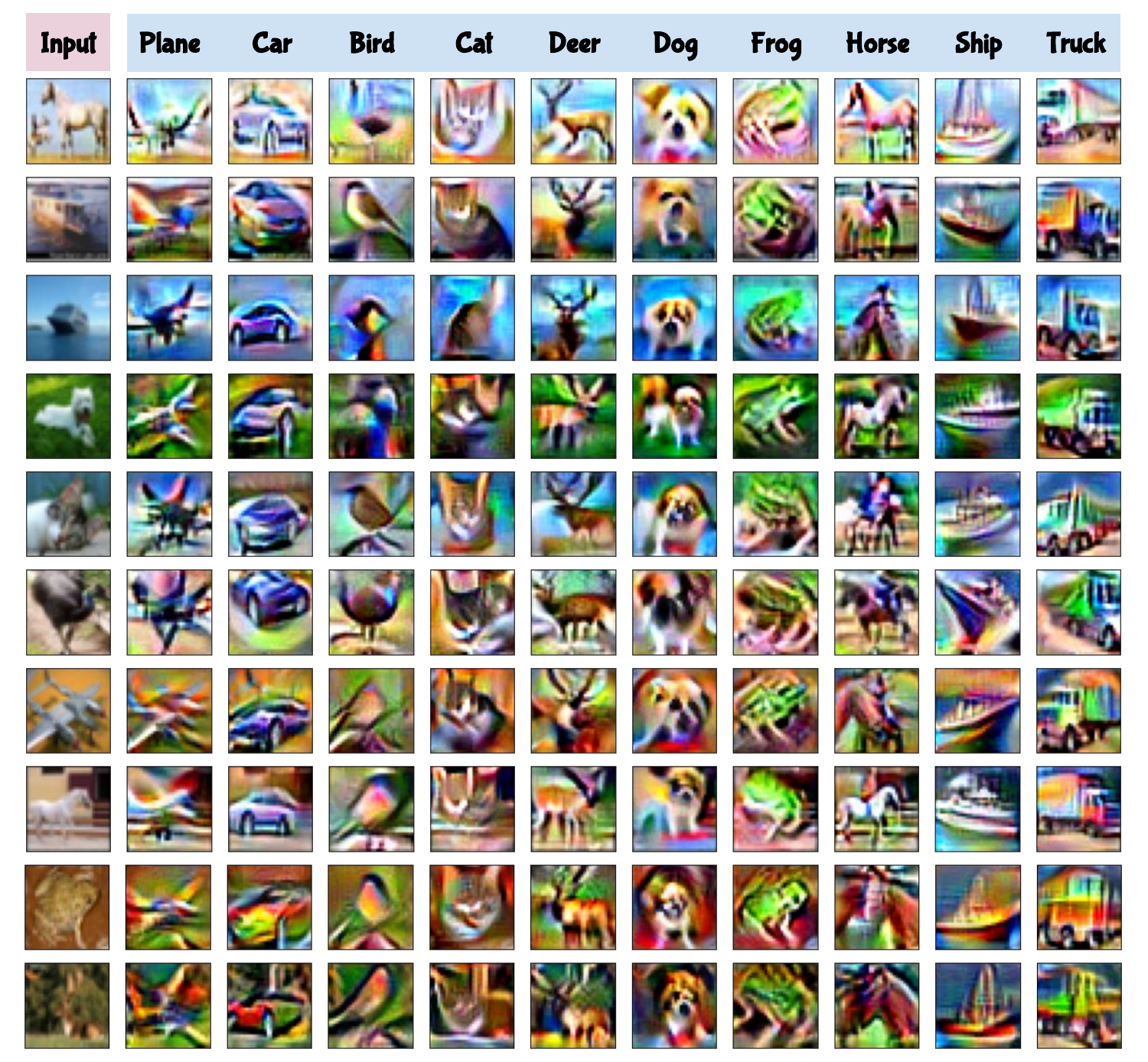}
    \caption{\textbf{CIFAR-10 PAG Visualizations of SBG}. Targeted large-$\epsilon$ adversarial examples on ResNet-18 trained on CIFAR-10 using SBG. We present 10 randomly selected images and transform them into all the possible classes. As can be seen, all images contain meaningful class-related semantic information.}
    \label{fig:c10_rn18_sbg}
\end{figure}

\textbf{ResNet-18}:
As can be seen in \cref{fig:rn_pag}, while ``vanilla'' classifiers do not possess PAG, both our approaches and AT models obtain aligned gradients. To better demonstrate the gradient obtained by SBG, we visualize in \cref{fig:c10_rn18_sbg} the output of applying strong targeted adversarial examples, starting from random images.

\end{document}